\documentclass{article}
\usepackage{ijcai20}

\usepackage{times}
\usepackage{soul}
\usepackage{url}
\usepackage[utf8]{inputenc}
\usepackage[small]{caption}
\usepackage{graphicx}
\usepackage{amsmath}
\usepackage{amsthm}
\usepackage{booktabs}
\usepackage{algorithm}
\usepackage{algorithmic}
\newcommand\ie{\textit{i.e.}}
\newcommand\eg{\textit{e.g.}}
\newcommand\etc{\textit{etc}}
\newcommand\wrt{\textit{w.r.t}}
\newcommand\unitrans{\text{UniTrans}}

\usepackage{bm, color, amssymb, latexsym, multirow}

\title{\unitrans{} : Unifying Model Transfer and Data Transfer for Cross-Lingual \\Named Entity Recognition with Unlabeled Data}

\author{
Qianhui Wu$^1$
\and
Zijia Lin$^2$\and
Börje F. Karlsson$^2$\and
Biqing Huang$^1$\And
Jian-Guang Lou$^2$
\affiliations
$^1$Beijing National Research Center for Information Science and Technology (BNRist)\\
Department of Automation, Tsinghua University, Beijing 100084, China\\
$^2$Microsoft Research, Beijing 100080, China\\
\emails
wuqianhui@tsinghua.org.cn,
\{zijlin,borje.karlsson,jlou\}@microsoft.com,
hbq@tsinghua.edu.cn
}

\begin{document}

\maketitle

\begin{abstract}
        Prior works in cross-lingual named entity recognition (NER) with no/little labeled data fall into two primary categories: model transfer based and data transfer based methods. 
        In this paper we find that both method types can complement each other, in the sense that, the former can exploit context information via language-independent features but sees no task-specific information in the target language; while the latter generally generates pseudo target-language training data via translation but its exploitation of context information is weakened by inaccurate translations.
        Moreover, prior works rarely leverage unlabeled data in the target language, which can be effortlessly collected and potentially contains valuable information for improved results.
        To handle both problems, we propose a novel approach termed \unitrans{} to Unify both model and data Transfer for cross-lingual NER, and furthermore, to leverage the available information from unlabeled target-language data via enhanced knowledge distillation.
        We evaluate our proposed \unitrans{} over 4 target languages on benchmark datasets.
        Our experimental results show that it substantially outperforms the existing state-of-the-art methods.
\end{abstract}

\section{Introduction}

Named entity recognition (NER) is a fundamental task in natural language processing, which seeks to locate and classify named entities, like locations, organizations, etc., in unstructured texts. NER has been extensively studied, especially monolingual NER, as it is widely incorporated in various downstream tasks, \eg{}, question answering~\cite{molla2006named}. 
Recently, deep neural networks have become the dominant approach due to their superior performance. 
However, the key to their success is the availability of adequate labeled training data; and building large labeled training sets for a new language of interest can be time consuming and labor costly.
This motivates researches on \emph{cross-lingual transfer}, which leverages labeled data from a source language (\eg, English) to overcome the data scarcity issue in the target language. 
In this paper, following~\cite{wu2019beto} and \cite{wu2020enhanced}, we focus on \emph{zero-resource cross-lingual transfer}, where \emph{NO labeled data} is available in the target language.
In this way, our work can act as a basis for further researches where some labeled target-language data exists. 

The state-of-the-art methods for zero-resource cross-lingual NER mainly fall into two categories: 
i) \emph{model transfer based methods}~\cite{wu2019beto,wu2020enhanced}, which use labeled source-language data to train an NER model with language-independent features (\eg{}, cross-lingual word representations~\cite{devlin2019bert}), and then directly apply it to the target language;
ii) \emph{data transfer based methods}, which generally construct pseudo labeled data in the target language via translating the source-language training data into parallel target-language data and mapping the entity labels, and then use such pseudo labeled data to train a target-language NER model.
For example, 
\cite{mayhew2017cheap} and \cite{xie2018neural} employed word-to-word and phrase-to-phrase translation, respectively, to generate target-language training data, so that entity labels in the source-language training data can be directly copied to the generated target-language data.

In this paper, we hold the idea that the aforementioned model transfer based methods and data transfer based methods are complementary to each other. 
Specifically, for the former, models trained with labeled source-language data only learn the knowledge \wrt. the task in the source language, but cannot see any task-specific information in the target language. 
And thus they probably rely more on context information mined by language independent features to make predictions, as word-label relations in the target language are unavailable for them.
As for the latter, though models have access to word-label pairs in the target language to exploit their relations, inaccurate translations caused by sense ambiguity and word order differences will probably weaken the models' capability to predict through context information. 
Moreover, both methods generally do not leverage the language information contained in the unlabeled target-language data to benefit the cross-lingual transfer.

Therefore, here we propose a novel approach termed \unitrans{}  to unify both model transfer and data transfer for cross-lingual NER, and furthermore, to leverage the beneficial information from unlabeled target-language data via enhanced knowledge distillation.
Specifically, following~\cite{lample2018word}, we first construct a pseudo training set for the target language by performing word-to-word translation and copying entity labels. 
Then, we use a pre-trained cross-lingual language model, \ie, multilingual BERT~\cite{devlin2019bert}, to generate language-independent features to train \emph{two} separate NER models. One is trained on the labeled source-language data, and the other is trained on the pseudo target-language training data. 
We then integrate the knowledge contained in the source-language training data, the pseudo target-language training data, and the unlabeled target-language data as follows: 
i) we fine-tune the model trained on source-language data with the pseudo target-language training data, and take the fine-tuned model as a teacher NER model to predict probability distributions of entity labels (\ie, soft labels) for each word in the unlabeled target-language data;
ii) we propose a voting scheme that associates the three aforementioned NER models to predict high-confidence one-hot label vectors (\ie{}, hard labels) for part of words in the unlabeled target-language data; 
iii) we train a student NER model on the unlabeled target-language data with supervision from both the aforementioned soft labels and hard labels, as how knowledge distillation works, and we use it as the final target-language NER model.

Extensive experiments conducted on benchmark datasets for 4 target languages well demonstrate that the proposed \unitrans{} substantially outperforms the existing state-of-the-art methods. We also extend \unitrans{} by ensembling multiple teacher models with different random seeds, and show further performance improvement.

Our major contributions are summarized as follows:
\begin{itemize}
    \item We propose a novel approach termed \unitrans{} to unify model transfer and data transfer based on their complementarity for cross-lingual NER, with the help of beneficial information from unlabeled target-language texts.
    \item We propose a voting scheme to generate pseudo hard labels on unlabeled target-language data, so as to enhance knowledge distillation in \unitrans{} with supervision from both soft labels and pseudo hard labels. 
    \item We conduct extensive experiments on benchmark datasets and show that \unitrans{} yields new state-of-the-art cross-lingual NER performance, which can even be promoted via teacher ensembling. 
\end{itemize}

\section{Related Work}
\subsection{Cross-Lingual NER}
Cross-lingual transfer for NER has attracted much attention in recent years. 
Prior works roughly fall into two primary categories: model transfer based and data transfer based.

Model transfer based methods generally utilize language-independent features to train an NER model on the labeled source-language data, so that the model can be directly applied to the target language. 
Those language-independent features include aligned word representations~\cite{ni2017weakly,wu2019beto}, word clusters~\cite{tackstrom2012}, Wikifier features~\cite{tsai2016cross}, and gazetteers~\cite{zirikly2015cross}, \etc. 
Besides directly applying the trained NER model,~\cite{wu2020enhanced} further proposed a meta-learning algorithm to improve the prediction for each test case with its similar examples in the labeled source-language data.

Data-transfer based methods generally train a monolingual NER model for the target language with pseudo training data constructed from the labeled source-language data. 
\cite{ni2017weakly} proposed to use bilingual parallel texts and their word alignment information to project labels from the source language to the target language. 
Considering it is expensive to build parallel corpora manually, most recent methods propose to translate the source-language texts into the target language in a word-by-word~\cite{xie2018neural} or phrase-by-phrase~\cite{mayhew2017cheap} manner, and then copy the label of each word/phrase to its corresponding translated word/phrase.
\cite{jain2019entity} further proposed to utilize Google Translate twice to translate sentences as well as entities, and align entity labels based on distributional statistics derived from the dataset.

As mentioned above, both model transfer and data transfer can be complementary to each other. 
And thus in this paper, we propose \unitrans{} to unify both, so as to retain the capability of predicting through context information and meanwhile exploiting word-label relations in the target language. 
Additionally, few prior works on cross-lingual NER leveraged unlabeled data in the target language. 
\cite{bari2019zero} proposed to fine-tune the model trained on source-language data with unlabeled target-language data in a manner similar to self-training~\cite{scudder1965probability}. 
As shown by \cite{he2017aunified} and \cite{bari2019zero} that unlabeled data is beneficial, our \unitrans{} also exploits the unlabeled target-language data via an enhanced knowledge distillation process, as mentioned before. 

\begin{figure*}[t]
    \centering
    \includegraphics[width=16.5cm]{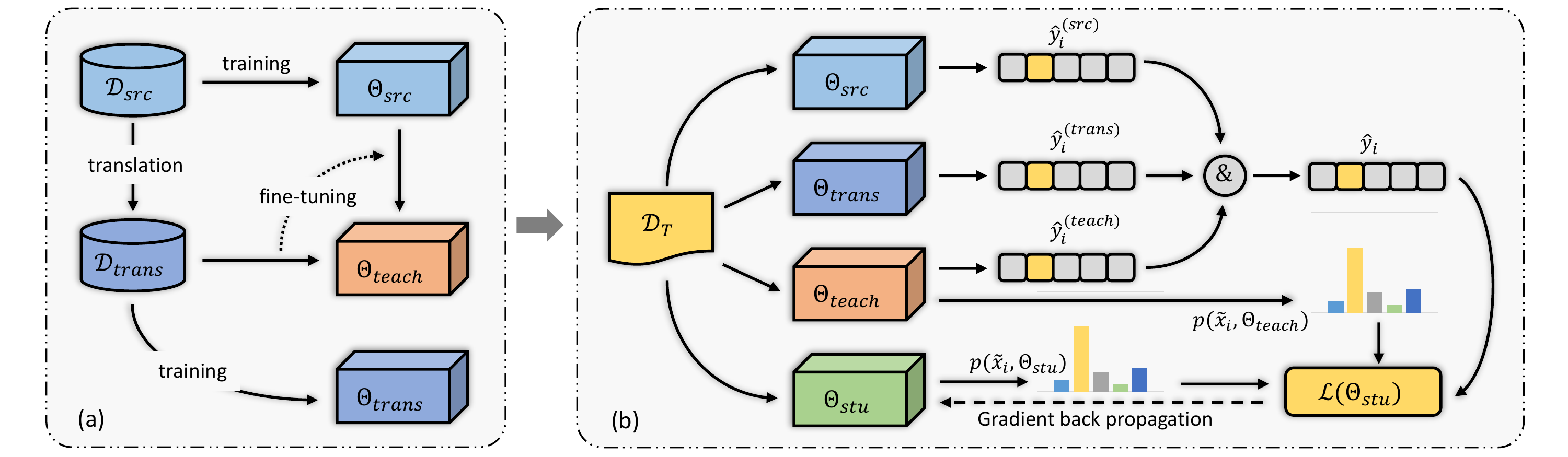}
    \caption{Framework of the proposed \unitrans{}. (a) Unifying model transfer and data transfer. (b) Knowledge distillation on unlabeled data.
    }
\label{fig:framework}
\end{figure*}

\subsection{Knowledge Distillation}
Knowledge distillation was originally proposed for model compression~\cite{bucilu2006model}, \ie, to learn a compact student model that retains most performance of a larger teacher model or ensemble of models that require more space to deploy or more computation to make predictions~\cite{rusu2015policy,hinton2015distilling,sanh2019distilbert}. 
Besides model compression, researchers also applied knowledge distillation to various tasks, like deep reinforcement learning~\cite{rusu2015policy}, image classification~\cite{hinton2015distilling}, and language modeling~\cite{sanh2019distilbert}.

In this paper, we adapt knowledge distillation to cross-lingual NER.
And different from typical application scenarios of knowledge distillation that do not consider unlabeled data, in our proposed \unitrans{}, the student model is completely trained on the unlabeled target-language data.
In addition, we propose a voting scheme to generate pseudo hard labels for the unlabeled target-language data, thus enhancing knowledge distillation with supervision from both soft labels and pseudo hard labels.

\section{Methodology}

Figure~\ref{fig:framework} illustrates the framework of the proposed \unitrans{}. 
Specifically, we first generate pseudo training data for the target language by performing word-to-word translation and copying entity labels. 
Then, we train an NER model using the labeled source-language data and derive a teacher model by fine-tuning it with the pseudo target-language training data. 
Finally, we adapt knowledge distillation to train a student NER model for the target language on unlabelled data.

\subsection{Data Transfer via Word-to-Word Translation}
\label{sec:data_trans}
Following~\cite{xie2018neural}, we apply ~\cite{lample2018word} to translate the source-language training data word-by-word into the target language, and then \textit{directly copy the entity label of each source-language word to its corresponding translated word}. 
The pipeline of data transfer is briefly introduced below. 
One can refer to~\cite{lample2018word} for more details.

We first leverage identical ``character strings''~\cite{smith2017offline} in both languages, \ie{}, shared words in most cases, to create a seed dictionary $\{s_i, t_i\}_{i=1}^D$, where $s_i,~t_i\in \mathbb{R}^d$, respectively, denote the source-language word embedding and the target-language word embedding of an identical character string, and $D$ is the size of the dictionary. 
Then, we learn a linear mapping $P\in \mathbb{R}^{d\times d}$ between the source-language embedding matrix $S = [s_1, s_2, ..., s_D]\in\mathbb{R}^{d\times D}$ and the target-language embedding matrix $T= [t_1, t_2, ..., t_D]\in\mathbb{R}^{d\times D}$, with its objective function formulated as follows. 
\begin{equation}
    \label{eq:Procrustes}
    P = \arg\min_{P'} \left\Vert P'S-T \right\Vert_F~~~~s.t.,~ P'^TP'=I
\end{equation}
where $\left\Vert \cdot \right\Vert_F$ denotes the Frobenius norm. 
The closed-form solution for Eq.~\ref{eq:Procrustes} can be derived via singular value decomposition (SVD):
\begin{equation}
    P = UV^T~~~~s.t.,~ U\Sigma V^T = \text{SVD}(TS^T)
\end{equation}

To produce word-to-word translations, for a source-language word, we use the learned $P$ to map its embedding vector into the target-language embedding space, and take its nearest neighbor in the target language as its corresponding translation result.
Specifically, we use the cross-domain similarity local scaling (CSLS)~\cite{lample2018word} to measure the distance between the mapped embedding vector of a source-language word (denoted as $Ps_i$) and the embedding vector of a target-language word (denoted as $t_j$):
\begin{equation}
    \label{fig:CSLS}
    \text{CSLS}(P s_i, t_j) = 2~cos(Ps_i, t_j) - r_T(Ps_i) - r_S(t_j)
\end{equation}
\begin{equation}
    r_T(Ps_i) = \frac{1}{K}\sum_{t_k\in \mathcal{N}_T(Ps_i)} cos(Ps_i, t_k)
\end{equation}
\begin{equation}
    r_S(t_j) = \frac{1}{K}\sum_{Ps_k\in \mathcal{N}_S(t_j)} cos(Ps_k, t_j)
\end{equation}
where $\mathcal{N}_{T}(Ps_i)$ denotes the $K$ target-language nearest neighbors of $Ps_i$, $\mathcal{N}_{S}(t_j)$ denotes the $K$ mapped source-language nearest neighbors of $t_j$, and $cos(\cdot)$ denotes cosine similarity. 


By word-by-word translation and copying word labels, we can build a pseudo target-language training set $\mathcal{D}_{trans}$.

\subsection{Base Model for NER}
\label{sec:base_model}
The base model for NER in this paper consists of a feature encoder and a linear classification layer. 
Given an input text sequence $\bm{x}=\{x_i\}_{i=1}^N$ with $N$ words, we first feed it into the feature encoder $f_{\theta}$ to obtain feature vectors $\bm{h}=\{h_i\}_{i=1}^N$ for all words:
\begin{equation}
    \label{eq:hidden}
    \bm{h} = f_{\theta}(\bm{x})
\end{equation}
where $f_{\theta}$ can be any feature encoder that produces language-independent features, and $h_i$ is the feature vector corresponding to the $i$-th word $x_i$. Following~\cite{wu2019beto}, here we utilize multilingual BERT~\cite{devlin2019bert} as the language-independent feature encoder.

Then for each word $x_i$, its corresponding feature vector $h_i$ is fed into the linear classification layer with the \emph{softmax} function to predict the probability distribution of entity labels for it, which is formulated as follows.
\begin{equation}
\label{eq:softmax}
    p(x_i, \Theta) = \text{softmax}(Wh_i + b)
\end{equation}
where $p(x_i, \Theta)\in \mathbb{R}^{|C|}$ with $C$ being the entity label set, and $\Theta=\{f_{\theta}, W, b\}$ denotes all the to-be-learned parameters.

Suppose a labeled training set is denoted as $\mathcal{D}=\{(\bm{x},\bm{y})\}$, where $\bm{y}=\{y_i\}_{i=1}^N$ is the corresponding one-hot entity label vectors for words in the corresponding $\bm{x}$. Then the loss function $\mathcal{L}$ for the NER model is defined as the cross entropy between the predicted probability distribution of entity labels and the ground-truth one for each word, as formulated below:
\begin{equation}
    \label{eq:loss_base}
    \mathcal{L}(\Theta) = \frac{1}{|\mathcal{D}|}\sum_{(\bm{x},\bm{y})\in \mathcal{D}} \frac{1}{N}\sum_{i=1}^N \text{CrossEntropy}
    \left({y_i, p(x_i, \Theta)}\right)
\end{equation}
where $y_i$ is the one-hot label vector corresponding to $x_i$.

Note that in this paper, we use the base model here to train all mentioned NER models.

\subsection{Unifying Model Transfer and Data Transfer}
\label{sec:unify}
With the base model and loss function defined above, we can train an NER model on the labeled source-language data like previous model transfer based methods~\cite{wu2019beto}, denoted as $\Theta_{src}$. Then for unifying model transfer and data transfer based on their complementarity, the proposed \unitrans{} further trains a teacher model $\Theta_{teach}$ by fine-tuning $\Theta_{src}$ on the pseudo target-language training data, with the same loss function as Eq.~\ref{eq:loss_base}. And thus $\Theta_{teach}$ is expected to combine the advantages of both model transfer and data transfer. Moreover, our experiments also show that, $\Theta_{teach}$ obtains superior performance than simply combining the labeled source-language data and the pseudo target-language training data to train an NER model.

\subsection{Knowledge Distillation on Unlabeled Data}
\label{sec:knowledge}

\subsubsection{Using Soft Labels}
For knowledge distillation, we first apply the teacher model $\Theta_{teach}$ to the unlabeled target-language data $\mathcal{D}_{T}=\{\bm{\tilde{x}}\}$, where $\bm{\tilde{x}}=\{\tilde{x}_i\}_i^N$ denotes an unlabeled target-language text with $N$ words. Considering that predicted soft labels (\ie{}, probability distribution of entity labels) can contain richer information than predicted hard labels \cite{hinton2015distilling}, we use the soft labels output by $\Theta_{teach}$ as the supervision to train a student model $\Theta_{stu}$, by minimizing the \emph{mean squared error} (MSE) between the prediction of $\Theta_{stu}$ and that of $\Theta_{teach}$ for each word. 
The loss function \wrt. an unlabelled target-language text $\bm{\tilde{x}}$ is formulated as:
\begin{equation}
    \label{eq:loss_soft}
    \mathcal{L}_{soft}^{\bm{\tilde{x}}} =  \frac{1}{N}\sum_{i=1}^N \text{MSE} \left( p(\tilde{x}_i, \Theta_{teach}), p(\tilde{x}_i, \Theta_{stu}) \right)
\end{equation}
where $p(\tilde{x}_i, \Theta_{stu})$ denotes the probability distribution of entity labels predicted by the student model for the $i$-th word $\tilde{x}_i$, and $p(\tilde{x}_i, \Theta_{teach})$ denotes that of the teacher model. 

In the way above, we can not only transfer the knowledge learned by the teacher model $\Theta_{teach}$ to the student model $\Theta_{stu}$, but also enable the student model $\Theta_{stu}$ to capture language specific information in unlabeled target-language data $\mathcal{D}_T$. And thus, with an identical base model, the student model $\Theta_{stu}$ is expected to gain performance improvement over the teacher model $\Theta_{teach}$.

\subsubsection{Incorporating Pseudo Hard Labels}
\cite{hinton2015distilling} pointed out that when correct hard labels are available, the student model trained with only soft labels can be further improved by also training it with supervision from the correct hard labels. 
However, in the case of zero-resource cross-lingual NER, there are no ground-truth entity labels available in $\mathcal{D}_T$. 
To this end, we propose a voting scheme to (partially) generate pseudo hard labels for the unlabeled target-language data. And we leverage them to help the training of the student model, via adding an extra loss to Eq.~\ref{eq:loss_soft}.

Specifically, we train another NER model in a supervised manner on the pseudo target-language training data generated by data transfer (section~\ref{sec:data_trans}), denoted as $\Theta_{trans}$. 
Then, we predict the probability distributions of entity labels for each word $\tilde{x}_i$ in any unlabeled text $\bm{\tilde{x}}$, using all learned NER models, \ie{}, $\Theta_{src}$, $\Theta_{teach}$ and $\Theta_{trans}$, respectively. 
And we take the entity label $c \in C$ with the highest probability as the predicted label $\hat{y}_i^{(*)}$ for $\tilde{x}_i$:
\begin{equation}
    \label{eq:inference_base}
    \hat{y}_i^{(*)} = \arg\max_c p(\tilde{x}_i, \Theta_{*})_c
\end{equation}
where $\Theta_{*}$ stands for $\Theta_{src}$, $\Theta_{teach}$ and $\Theta_{trans}$, with the corresponding $\hat{y}_i^{(*)}$ being $\hat{y}_i^{(src)}$, $\hat{y}_i^{(teach)}$ and $\hat{y}_i^{(trans)}$.
And we only generate a pseudo hard label $\hat{y}_i=\hat{y}_i^{(teach)}$ for the word $\hat{x}_i$ when $\hat{y}_i^{(teach)}=\hat{y}_i^{(src)}=\hat{y}_i^{(trans)}$, as $\hat{y}_i$ would be of high-confidence then. Here $\hat{y}_i$ can be seen as the voting result of $\Theta_{src}$, $\Theta_{teach}$ and $\Theta_{trans}$.
We denote the set of such words with a pseudo hard label as $\mathcal{X}$:
\begin{equation}
    \mathcal{X}=\{\tilde{x}_i| \hat{y}_i^{(teach)}=\hat{y}_i^{(src)}=\hat{y}_i^{(trans)},~\forall \tilde{x}_i \in \mathcal{D}_T\}
\end{equation}

With the pseudo hard labels derived, we further define a hard label based loss $\mathcal{L}_{hard}^{\bm{\tilde{x}}}$ as follows for the unlabeled target-language text $\bm{\tilde{x}}$.
\begin{equation}
    \mathcal{L}_{hard}^{\bm{\tilde{x}}} = \frac{1}{N}\sum_{i=1}^N I(\tilde{x}_i)\cdot \text{CrossEntropy}(\hat{y}_i, p(\tilde{x}_i, \Theta_{stu}))
\end{equation}
where $I(\tilde{x}_i)$ is an indicator function that returns $1$ if $\tilde{x}_i \in \mathcal{X}$ and otherwise $0$. 

Then the loss function combining both the supervision from pseudo hard labels and that from the soft labels output by $\Theta_{teach}$ is formulated as follows to better guide the training of $\Theta_{stu}$:
\begin{equation}
    \label{eq:loss_final}
    \mathcal{L}(\Theta_{stu}) = \frac{1}{\vert\mathcal{D}_T\vert}\sum_{\bm{\tilde{x}} \in \mathcal{D}_T} \left(\eta\mathcal{L}_{hard}^{\bm{\tilde{x}}} + \mathcal{L}_{soft}^{\bm{\tilde{x}}}\right)
\end{equation}
where $\eta > 0$ is a trade-off parameter, and we simply set $\eta = 1$ in this paper. With $\mathcal{L}(\Theta_{stu})$, the student model $\Theta_{stu}$ is learned in an end-to-end manner.


\subsection{Inference in Target Language}
For inference in the target language, we only use the learned student model $\Theta_{stu}$ to predict the probability distribution of entity labels for each word in a given test case, as Eq.~\ref{eq:softmax}. 
Note that here we do not straightly take the entity label with the highest probability as the prediction result for each word (Eq.~\ref{eq:inference_base}). Instead, we apply \emph{Viterbi} decoding~\cite{chen2019grn} by just adding constraints to ensure that predicted entity labels for all words in the test case would not violate the NER tagging scheme and meanwhile would obtain the highest probability as a label sequence. And thus we don't need to train a transition matrix here.

\section{Experiments}
We evaluate the proposed \unitrans{} for zero-resource cross-lingual NER through experiments over benchmark datasets on 4 target languages (\ie, Spanish, Dutch, German, and Norwegian\footnote{We use Bokmål rather than Nynorsk here, considering that it is used by 85–90\% of the population of Norway.}) and comparisons with state-of-the-art methods. 

\begin{table}[t]
    \centering
    \scalebox{0.8}{
    \setlength{\tabcolsep}{1.5mm}
    \begin{tabular}{c|c|c|c|c}
        \toprule
        Language & Type &Train &Dev &Test  \\ \midrule
        English [en] & \# of Sentence &14,987 &3,466 &3,684 \\
        (CoNLL-2003)& \# of Entity &23,499 &5,942 &5,648 \\
        \midrule
        German [de] & \# of Sentence &12,705 &3,068 &3,160 \\
        (CoNLL-2003)& \# of Entity &11,851 &4,833 &3,673 \\
        \midrule
        Spanish [es] & \# of Sentence &8,323 &1,915 &1,517 \\
        (CoNLL-2002)& \# of Entity &18,798 &4,351 &3,558 \\
        \midrule
        Dutch [nl] & \# of Sentence &15,806 &2,895 &5,195 \\
        (CoNLL-2002)& \# of Entity &13,344 &2,616 &3,941 \\
        \midrule
        Norwegian [no] & \# of Sentence &15,686 &2,410 &1,939 \\
        (NoDaLiDa-2019)& \# of Entity &10,934 &1,615 &1,391 \\
        \toprule
    \end{tabular}
    }
    \caption{Statistics of the benchmark datasets.}
    \label{tab:dataset}
\end{table}

\subsection{Experiment Settings}
\subsubsection{Datasets}
We use the following widely-used benchmark datasets for experiments:
CoNLL-2002~\cite{tjong2002introduction} for Spanish [es] and Dutch [nl] NER, CoNLL-2003~\cite{tjong2003introduction} for English [en] and German [de] NER, and NoDaLiDa-2019~\cite{johansen2019ner} for Norwegian [no] NER. 
All datasets are annotated with 4 entity types: \texttt{LOC}, \texttt{MISC}, \texttt{ORG}, and \texttt{PER}. 
Each dataset is split into training, dev, and test sets.
Table \ref{tab:dataset} reports the statistics of each.

We leverage WordPiece~\cite{wu2016google} to tokenize each sentence into a sequence of subwords and, following~\cite{wu2019beto,wu2020enhanced}, we use the BIO entity labeling scheme.
Moreover, as previous works~\cite{wu2020enhanced}, all experiments use English as the source language and the others as the target language. 

Note that for each target language, we delete all entity labels in its training set, and use it only as unlabeled target-language data. Moreover, to imitate the zero-resource cross-lingual NER case, we ignore all target-language dev sets, and directly evaluate the learned models on their test sets.

\subsubsection{Implementation Details}
We implement our \unitrans{} with PyTorch\footnote{https://pytorch.org/}.
For word-to-word translation, we use \emph{fastText}\footnote{https://fasttext.cc/docs/en/pretrained-vectors.html} monolingual word embeddings, and use \emph{MUSE}\footnote{https://github.com/facebookresearch/MUSE}~\cite{lample2018word} to perform translation. 
For the feature encoder of the base model, \ie{}, $f_{\theta}$ in Eq.~\ref{eq:hidden}, we employ the pretrained multilingual BERT model (case-sensitive version)  
~\cite{devlin2019bert} in \emph{HuggingFace's Transformers}\footnote{https://github.com/huggingface/transformers}, which has 12 Transformer blocks, 12 attention heads, and 768 hidden units. 

We empirically set \unitrans{} hyper-parameters by following previous works (as cited below), and utilize them in all experiments. 
Specifically, we adopt a dropout rate of 0.1~\cite{wu2020enhanced} and freeze the parameters of the embedding layer and the bottom three layers of the multilingual BERT~\cite{wu2019beto}.
Following~\cite{wu2020enhanced}, we train all models for 3 epochs using a batch size of 32, maximum sequence length of 128, and AdamW~\cite{loshchilov2017fixing} as the optimizer. For AdamW, we use a learning rate of $5\mathrm{e}{-5}$ ~\cite{wolf2019transformers} for teacher models and $1\mathrm{e}{-4}$ for the student model~\cite{yang2019model}.
Note that if a word is split into several subwords after tokenization, only the first subword is considered in the loss function.

\subsubsection{Performance Metric}
Following~\cite{tjong2002introduction}, we use entity level F1-score as the performance metric. 
Moreover, we conduct each experiment 5 times and report the mean F1-score.

\begin{table}[t]
    \centering  
    \setlength{\tabcolsep}{0.8mm}
    \scalebox{0.8}{
    \begin{tabular}{c|ccccc}
        \toprule
        & es & nl & de & no & Average \\ \midrule
        \citeauthor{tackstrom2012}~\shortcite{tackstrom2012}    & 59.30 & 58.40 & 40.40 & - & - \\
        \citeauthor{tsai2016cross}~\shortcite{tsai2016cross}    & 60.55 & 61.56 & 48.12 & - & - \\
        \citeauthor{ni2017weakly}~\shortcite{ni2017weakly}      & 65.10 & 65.40 & 58.50 & - & -\\ 
        \citeauthor{mayhew2017cheap}~\shortcite{mayhew2017cheap}& 64.10 & 63.37 & 57.23 & - & -\\ 
        \citeauthor{xie2018neural}~\shortcite{xie2018neural}    & 72.37 & 71.25 & 57.76 & - & -\\ 
        \citeauthor{jain2019entity}~\shortcite{jain2019entity}  & 73.5  & 69.9  & 61.5  & - & -\\ 
        \citeauthor{bari2019zero}~\shortcite{bari2019zero}      & 75.93 & 74.61 & 65.24 & - & -\\ 
        \citeauthor{wu2019beto}~\shortcite{wu2019beto}$^{\dag}$ & 74.50 & 79.50 & 71.10 & - & -\\ 
        \citeauthor{wu2020enhanced}~\shortcite{wu2020enhanced}  & 76.75 & 80.44 & 73.16 & - & -\\ 
        \midrule
        \multirow{2}{*}{Model Transfer (reimp.)}
        & 76.34 & 80.61 & 72.39 & 78.47 & \multirow{2}{*}{76.95}\\ 
        & ($\pm$ 0.96) & ($\pm$ 0.46) & ($\pm$ 1.05) & ($\pm$ 0.36) &\\
        \multirow{2}{*}{Data Transfer (reimp.)}
        & 78.14 & 80.98 & 73.65 & 78.91 & \multirow{2}{*}{77.92}\\ 
        & ($\pm$ 0.97) & ($\pm$ 0.72) & ($\pm$ 0.36) & ($\pm$ 0.50) &\\
        \multirow{2}{*}{\textbf{\unitrans{}}}
        & \textbf{79.31} & \textbf{82.90} & \textbf{74.82} & \textbf{81.17} & \multirow{2}{*}{\textbf{79.55}}\\
        & ($\pm$ 0.39) & ($\pm$ 0.43) & ($\pm$ 0.60) & ($\pm$ 0.63) &\\
        \bottomrule
    \end{tabular}
    }
    \caption{Results of the proposed \unitrans{} and prior state-of-the-art methods for zero-resource cross-lingual NER. $^{\dag}$ denotes the reported results \wrt. freezing the bottom 3 layers of BERT as in this paper. We also report the standard deviation for reimplemented baselines and the proposed \unitrans{} (\ie, numbers in parentheses).
    }
    \label{tab:results}
\end{table}

\begin{table*}[t]
    \centering
    \scalebox{0.76}{
    \begin{tabular}{l|ccccc}
        \toprule
        &   es  &   nl  &   de  &   no  &   Average \\
        \midrule
        \textbf{\unitrans{}} &  79.31   &   82.90   &   74.82   &   81.17   &   79.55   \\ \midrule
        1) \unitrans{} w/o $\mathcal{L}_{soft}^{\bm{\tilde{x}}}$ &
            78.93 (-0.38)   &   82.48 (-0.42)   &   75.23 (0.41)    &   80.91 (-0.26)   &   79.39 (-0.16)\\ 
        2) \unitrans{} w/o $\mathcal{L}_{hard}^{\bm{\tilde{x}}}$ &
            79.54 (0.23)    &   82.78 (-0.12)   &   74.43 (-0.39)   &   81.06 (-0.11)   &   79.45 (-0.10)\\
        \midrule
        3) \unitrans{} w/ $\Theta_{src}$  &
            77.30 (-2.01)   &   81.20 (-1.70)   &   73.61 (-1.21)   &   80.42 (-0.75)   &   78.13 (-1.42)\\ 
        4) \unitrans{} w/ $\Theta_{trans}$ &
            79.24 (-0.07)   &   82.13 (-0.77)   &   74.91 (0.09)    &   80.06 (-1.11)   &   79.09 (-0.46)\\ 
        5) \unitrans{} w/o $\mathcal{D}_T$ (\ie{}, $\Theta_{teach}$) &
            78.24 (-1.07)   &   81.73 (-1.17)   &   73.97 (-0.85)   &   79.07 (-2.10)   &   78.25 (-1.30) \\
        \midrule
        6) Model Transfer (\ie{}, $\Theta_{src}$) &
            76.34 (-2.97)   &   80.61 (-2.29)   &   72.39 (-2.43)   &   78.47 (-2.70)   & 76.95 (-2.60)\\
        7) Data Transfer (\ie{}, $\Theta_{trans}$) &
            78.14 (-1.17)   &   80.98 (-1.92)   &   73.65 (-1.17)   &   78.91 (-2.26)   &   77.92 (-1.63) \\
        8) Data Combination &
            77.31 (-2.00)   &   80.75 (-2.15)   &   73.66 (-1.16)   &   76.59 (-4.58)   &   77.08 (-2.47) \\
        \bottomrule
    \end{tabular}
    }
    \caption{Ablation study for the proposed \unitrans{}, where numbers in parenthesis denote performance changes. 
    } 
    \label{tab:ablation}
\end{table*}

\subsection{Experimental Results}
Table~\ref{tab:results} reports the zero-resource cross-lingual NER results of the proposed \unitrans{} on the 4 target languages, alongside those reported by prior state-of-the-art methods and those of two re-implemented baseline methods, \ie, \emph{Model Transfer} ($\Theta_{src}$ in~\ref{sec:unify}) and \emph{Data Transfer} ($\Theta_{trans}$ in~\ref{sec:knowledge}). 

Table~\ref{tab:results} shows that our proposed \unitrans{} significantly outperforms the prior state-of-the-art methods and the re-implemented baselines on all target languages. 
Particularly, compared with the best prior method~\cite{wu2020enhanced}, the proposed \unitrans{} achieves an improvement of F1-score ranging from $1.66$ for German [de] to $2.56$ for Spanish [es].
Moreover, \unitrans{} achieves an average improvement of $1.63$ F1-score over \emph{Data Transfer},  
and $2.60$ F1-score over \emph{Model Transfer}.
All these results well demonstrate the effectiveness of the proposed \unitrans{}, which is mainly attributed to unifying model transfer with data transfer and leveraging unlabeled target-language data in \unitrans{}.


\subsection{Ablation Study}
To validate the contributions of different components in the proposed \unitrans{}, we introduce the following variants of \unitrans{} and baselines to perform ablation study: 
1) \emph{\unitrans{} w/o $\mathcal{L}_{soft}^{\bm{\tilde{x}}}$}, which removes $\mathcal{L}_{soft}^{\bm{\tilde{x}}}$ from Eq.\ref{eq:loss_final} to train the target-language NER model $\Theta_{stu}$ with only the pseudo hard labels; 
2) \emph{\unitrans{} w/o $\mathcal{L}_{hard}^{\bm{\tilde{x}}}$}, which eliminates $\mathcal{L}_{hard}^{\bm{\tilde{x}}}$ from Eq.\ref{eq:loss_final} (\ie{}, $\eta=0$) to train $\Theta_{stu}$ with only soft labels output by the teacher model $\Theta_{teach}$; 
3) \emph{\unitrans{} w/ $\Theta_{src}$}, which uses the model $\Theta_{src}$ learned on the source-language training data with language-independent features as the teacher to train $\Theta_{stu}$ on unlabeled target-language data;
4) \emph{\unitrans{} w/ $\Theta_{trans}$}, which uses the model $\Theta_{trans}$ learned on translated target-language training data as the teacher to train $\Theta_{stu}$ on unlabeled target-language data; 
5) \emph{\unitrans{} w/o $\mathcal{D}_{T}$} (\ie{}, $\Theta_{teach}$), which cuts out the access to the unlabeled target-language data $\mathcal{D}_{T}$ and directly applies the teacher model $\Theta_{teach}$ to the target language;
6) \emph{Model Transfer} (\ie{}, $\Theta_{src}$), which directly applies the model $\Theta_{src}$ to the target language;
7) \emph{Data Transfer} (\ie{}, $\Theta_{trans}$), which directly applies the model $\Theta_{trans}$ to the target language;
8) \emph{Data Combination}, which directly combines the labeled source-language data and the translated target-language data to train an NER model for the target language with language-independent features.
Note that for variants 3) \emph{\unitrans{} w/ $\Theta_{trans}$} and 4) \emph{\unitrans{} w/ $\Theta_{src}$}, the hard label loss $\mathcal{L}_{hard}^{\bm{\tilde{x}}}$ is also removed from Eq.~\ref{eq:loss_final}, as we cannot use one model to generate pseudo hard labels.

Table~\ref{tab:ablation} highlights the performance contributions of each component in our proposed \unitrans{}, and removing any of them will generally lead to a performance drop. 
Moreover, we can draw more in-depth observations as follows.

1) The proposed \unitrans{} outperforms \emph{\unitrans{} w/o $\mathcal{L}_{soft}^{\bm{\tilde{x}}}$} and \emph{\unitrans{} w/o $\mathcal{L}_{hard}^{\bm{\tilde{x}}}$} in most cases, indicating that combining both for training the target-language model $\Theta_{stu}$ is reasonable. That also validates the effectiveness of the proposed voting scheme to generate pseudo hard labels. 

2) \emph{\unitrans{} w/o $\mathcal{L}_{hard}^{\bm{\tilde{x}}}$} outperforms \emph{\unitrans{} w/ $\Theta_{trans}$} and \emph{\unitrans{} w/ $\Theta_{src}$}. Meanwhile, \emph{\unitrans{} w/o $\mathcal{D}_{T}$} (\ie{}, $\Theta_{teach}$) outperforms \emph{Model Transfer} (\ie{}, $\Theta_{src}$) and \emph{Data Transfer} (\ie{}, $\Theta_{trans}$). Such results well demonstrate that the proposed teacher model $\Theta_{teach}$, which unifies model transfer and data transfer, is superior to just using one or the other,
no matter whether unlabelled target-language data is used or not.
That also well verifies the complementarity between model transfer and data transfer.

3) Comparing \unitrans{} with \emph{\unitrans{} w/o $\mathcal{D}_{T}$} (\ie{}, $\Theta_{teach}$), \emph{\unitrans{} w/ $\Theta_{src}$} with \emph{Model Transfer} (\ie{}, $\Theta_{src}$), and \emph{\unitrans{} w/ $\Theta_{trans}$} with \emph{Data Transfer} (\ie{}, $\Theta_{trans}$), respectively, we can see that eliminating the usage of unlabeled target-language data will lead to a consistent performance drop in all experiments. That further demonstrates the importance of leveraging information contained in unlabeled target-language data for cross-lingual NER. 

4) \emph{\unitrans{} w/o $\mathcal{D}_{T}$} (\ie{}, $\Theta_{teach}$) outperforms \emph{Data Combination}, meaning that the proposed $\Theta_{teach}$ utilizes a superior way to unify model transfer and data transfer than simply combining available labeled data.

\begin{table}[t]
    \centering
    \scalebox{0.77}{
    \setlength{\tabcolsep}{1.3mm}
    \begin{tabular}{l|cccccc}
        \toprule
        &   es  &   nl  &   de  &   no  &   Average \\ \midrule
        \unitrans{}                 &   \textbf{79.31}  &   82.90   &   74.82   &   81.17   &   79.55\\
        + Teacher Ensembling &  79.26   &   \textbf{83.07}  &   \textbf{75.55}  &   \textbf{81.30}  &   \textbf{79.79}\\ 
        \bottomrule
    \end{tabular}
    }
    \caption{Results of teacher ensembling ($M=5$) for \unitrans{}} 
    \label{tab:teascher_ensemble}
\end{table}
\subsection{Discussion: Extend with Teacher Ensembling}
Considering the randomness brought by dropout layers in the feature encoder (\ie{}, multilingual BERT) of the proposed \unitrans{}, we can extend \unitrans{} with teacher ensembling via taking advantage of the randomness. 
Specifically, we use an ensemble of teacher models learned via $M$ runs, each denoted as $\Theta^{(m)}_{*} \in \{\Theta^{(m)}_{src}, \Theta^{(m)}_{teach}, \Theta^{(m)}_{trans}\}$ with $m=1, 2, ..., M$. And we simply average the predictions of $\Theta^{(m)}_{*}$ to train the student target-language model $\Theta_{stu}$:
\begin{equation}
    \label{eq:ensemble}
    p(\tilde{x}_i, \Theta_{*}) = \frac{1}{M} \sum_{m=1}^M p(\tilde{x}_i, \Theta^{(m)}_{*})
\end{equation}
where $\Theta_* \in\{\Theta_{src}, \Theta_{teach}, \Theta_{trans}\}$.
Note that Eq.~\ref{eq:ensemble} is not only adopted to predict soft labels for unlabeled target-language data, but also employed in the generation of pseudo hard labels.

Table~\ref{tab:teascher_ensemble} reports the results of teacher ensembling with $M=5$. It is evident that ensembling further brings consistent performance improvements on nearly all target languages for \unitrans{}, with an average gain of 0.24 F1-score.  

\section{Conclusion}
In this paper, we propose a novel approach for cross-lingual NER termed \unitrans{}, which unifies both model transfer and data transfer based on their complementarity via enhanced knowledge distillation on unlabeled target-language data.
We also propose a voting scheme to generate pseudo hard labels for part of words in the unlabeled target-language data, so as to enhance knowledge distillation with supervision from both hard and soft labels.
We evaluate the proposed \unitrans{} on benchmark datasets for four target languages. Experimental results show that \unitrans{} achieves new state-of-the-art performance for all target languages. 
We also extend \unitrans{} with teacher ensembling, which leads to further performance gains.

\bibliographystyle{named}
\bibliography{ijcai20}

\begin{thebibliography}{}

\bibitem[\protect\citeauthoryear{Bari \bgroup \em et al.\egroup
  }{2019}]{bari2019zero}
M~Saiful Bari, Shafiq Joty, and Prathyusha Jwalapuram.
\newblock Zero-resource cross-lingual named entity recognition.
\newblock {\em arXiv preprint arXiv:1911.09812}, 2019.

\bibitem[\protect\citeauthoryear{Bucilu \bgroup \em et al.\egroup
  }{2006}]{bucilu2006model}
Cristian Bucilu, Rich Caruana, and Alexandru Niculescu-Mizil.
\newblock Model compression.
\newblock In {\em SIGKDD}, pages 535--541, 2006.

\bibitem[\protect\citeauthoryear{Chen \bgroup \em et al.\egroup
  }{2019}]{chen2019grn}
Hui Chen, Zijia Lin, Guiguang Ding, Jianguang Lou, Yusen Zhang, and Borje
  Karlsson.
\newblock Grn: Gated relation network to enhance convolutional neural network
  for named entity recognition.
\newblock In {\em Proceedings of the AAAI Conference on Artificial
  Intelligence}, pages 6236--6243, 2019.

\bibitem[\protect\citeauthoryear{Devlin \bgroup \em et al.\egroup
  }{2019}]{devlin2019bert}
Jacob Devlin, Ming-Wei Chang, Kenton Lee, and Kristina Toutanova.
\newblock {BERT}: Pre-training of deep bidirectional transformers for language
  understanding.
\newblock In {\em NAACL-HLT}, pages 4171--4186, 2019.

\bibitem[\protect\citeauthoryear{He and Sun}{2017}]{he2017aunified}
Hangfeng He and Xu~Sun.
\newblock A unified model for cross-domain and semi-supervised named entity
  recognition in chinese social media.
\newblock In {\em AAAI}, 2017.

\bibitem[\protect\citeauthoryear{Hinton \bgroup \em et al.\egroup
  }{2015}]{hinton2015distilling}
Geoffrey Hinton, Oriol Vinyals, and Jeff Dean.
\newblock Distilling the knowledge in a neural network.
\newblock {\em arXiv preprint arXiv:1503.02531}, 2015.

\bibitem[\protect\citeauthoryear{Jain \bgroup \em et al.\egroup
  }{2019}]{jain2019entity}
Alankar Jain, Bhargavi Paranjape, and Zachary~C. Lipton.
\newblock Entity projection via machine translation for cross-lingual {NER}.
\newblock In {\em EMNLP}, pages 1083--1092, 2019.

\bibitem[\protect\citeauthoryear{Johansen}{2019}]{johansen2019ner}
Bjarte Johansen.
\newblock Named-entity recognition for norwegian.
\newblock In {\em NoDaLiDa}, 2019.

\bibitem[\protect\citeauthoryear{Lample \bgroup \em et al.\egroup
  }{2018}]{lample2018word}
Guillaume Lample, Alexis Conneau, Marc'Aurelio Ranzato, Ludovic Denoyer, and
  Hervé Jégou.
\newblock Word translation without parallel data.
\newblock In {\em ICLR}, 2018.

\bibitem[\protect\citeauthoryear{Loshchilov and
  Hutter}{2017}]{loshchilov2017fixing}
Ilya Loshchilov and Frank Hutter.
\newblock Fixing weight decay regularization in adam.
\newblock {\em arXiv preprint arXiv:1711.05101}, 2017.

\bibitem[\protect\citeauthoryear{Mayhew \bgroup \em et al.\egroup
  }{2017}]{mayhew2017cheap}
Stephen Mayhew, Chen-Tse Tsai, and Dan Roth.
\newblock Cheap translation for cross-lingual named entity recognition.
\newblock In {\em EMNLP}, pages 2536--2545, 2017.

\bibitem[\protect\citeauthoryear{Moll{\'a} \bgroup \em et al.\egroup
  }{2006}]{molla2006named}
Diego Moll{\'a}, Menno van Zaanen, and Daniel Smith.
\newblock Named entity recognition for question answering.
\newblock In {\em ALTA}, pages 51--58, 2006.

\bibitem[\protect\citeauthoryear{Ni \bgroup \em et al.\egroup
  }{2017}]{ni2017weakly}
Jian Ni, Georgiana Dinu, and Radu Florian.
\newblock Weakly supervised cross-lingual named entity recognition via
  effective annotation and representation projection.
\newblock In {\em ACL}, pages 1470--1480, 2017.

\bibitem[\protect\citeauthoryear{Rusu \bgroup \em et al.\egroup
  }{2015}]{rusu2015policy}
Andrei~A. Rusu, Sergio~Gomez Colmenarejo, Caglar Gulcehre, Guillaume
  Desjardins, James Kirkpatrick, Razvan Pascanu, Volodymyr Mnih, Koray
  Kavukcuoglu, and Raia Hadsell.
\newblock Policy distillation.
\newblock {\em arXiv preprint arXiv:1511.06295}, 2015.

\bibitem[\protect\citeauthoryear{Sanh \bgroup \em et al.\egroup
  }{2019}]{sanh2019distilbert}
Victor Sanh, Lysandre Debut, Julien Chaumond, and Thomas Wolf.
\newblock Distilbert, a distilled version of bert: smaller, faster, cheaper and
  lighter.
\newblock {\em arXiv preprint arXiv:1910.01108}, 2019.

\bibitem[\protect\citeauthoryear{Scudder}{1965}]{scudder1965probability}
H~Scudder.
\newblock Probability of error of some adaptive pattern-recognition machines.
\newblock {\em IEEE Transactions on Information Theory}, pages 363--371, 1965.

\bibitem[\protect\citeauthoryear{Smith \bgroup \em et al.\egroup
  }{2017}]{smith2017offline}
Samuel~L Smith, David~HP Turban, Steven Hamblin, and Nils~Y Hammerla.
\newblock Offline bilingual word vectors, orthogonal transformations and the
  inverted softmax.
\newblock {\em arXiv preprint arXiv:1702.03859}, 2017.

\bibitem[\protect\citeauthoryear{T{\"a}ckstr{\"o}m \bgroup \em et al.\egroup
  }{2012}]{tackstrom2012}
Oscar T{\"a}ckstr{\"o}m, Ryan McDonald, and Jakob Uszkoreit.
\newblock Cross-lingual word clusters for direct transfer of linguistic
  structure.
\newblock In {\em NAACL}, pages 477--487, 2012.

\bibitem[\protect\citeauthoryear{Tjong Kim~Sang and
  De~Meulder}{2003}]{tjong2003introduction}
Erik~F. Tjong Kim~Sang and Fien De~Meulder.
\newblock Introduction to the {C}o{NLL}-2003 shared task: Language-independent
  named entity recognition.
\newblock In {\em HLT-NAACL}, pages 142--147, 2003.

\bibitem[\protect\citeauthoryear{Tjong Kim~Sang}{2002}]{tjong2002introduction}
Erik~F. Tjong Kim~Sang.
\newblock Introduction to the {C}o{NLL}-2002 shared task: Language-independent
  named entity recognition.
\newblock In {\em COLING}, 2002.

\bibitem[\protect\citeauthoryear{Tsai \bgroup \em et al.\egroup
  }{2016}]{tsai2016cross}
Chen-Tse Tsai, Stephen Mayhew, and Dan Roth.
\newblock Cross-lingual named entity recognition via wikification.
\newblock In {\em CoNLL}, pages 219--228, 2016.

\bibitem[\protect\citeauthoryear{Wolf \bgroup \em et al.\egroup
  }{2019}]{wolf2019transformers}
Thomas Wolf, Lysandre Debut, Victor Sanh, Julien Chaumond, Clement Delangue,
  Anthony Moi, Pierric Cistac, Tim Rault, R{\'e}mi Louf, Morgan Funtowicz,
  et~al.
\newblock Transformers: State-of-the-art natural language processing.
\newblock {\em arXiv preprint arXiv:1910.03771}, 2019.

\bibitem[\protect\citeauthoryear{Wu and Dredze}{2019}]{wu2019beto}
Shijie Wu and Mark Dredze.
\newblock Beto, bentz, becas: The surprising cross-lingual effectiveness of
  {BERT}.
\newblock {\em arXiv preprint arXiv:1904.09077}, 2019.

\bibitem[\protect\citeauthoryear{Wu \bgroup \em et al.\egroup
  }{2016}]{wu2016google}
Yonghui Wu, Mike Schuster, Zhifeng Chen, Quoc~V Le, Mohammad Norouzi, Wolfgang
  Macherey, Maxim Krikun, Yuan Cao, Qin Gao, Klaus Macherey, et~al.
\newblock Google's neural machine translation system: Bridging the gap between
  human and machine translation.
\newblock {\em arXiv preprint arXiv:1609.08144}, 2016.

\bibitem[\protect\citeauthoryear{Wu \bgroup \em et al.\egroup
  }{2020}]{wu2020enhanced}
Qianhui Wu, Zijia Lin, Guoxin Wang, Hui Chen, B{\"o}rje~F Karlsson, Biqing
  Huang, and Chin-Yew Lin.
\newblock Enhanced meta-learning for cross-lingual named entity recognition
  with minimal resources.
\newblock In {\em AAAI}, 2020.

\bibitem[\protect\citeauthoryear{Xie \bgroup \em et al.\egroup
  }{2018}]{xie2018neural}
Jiateng Xie, Zhilin Yang, Graham Neubig, Noah~A. Smith, and Jaime Carbonell.
\newblock Neural cross-lingual named entity recognition with minimal resources.
\newblock In {\em EMNLP}, pages 369--379, 2018.

\bibitem[\protect\citeauthoryear{Yang \bgroup \em et al.\egroup
  }{2019}]{yang2019model}
Ze~Yang, Linjun Shou, Ming Gong, Wutao Lin, and Daxin Jiang.
\newblock Model compression with two-stage multi-teacher knowledge distillation
  for web question answering system.
\newblock {\em arXiv preprint arXiv:1910.08381}, 2019.

\bibitem[\protect\citeauthoryear{Zirikly and Hagiwara}{2015}]{zirikly2015cross}
Ayah Zirikly and Masato Hagiwara.
\newblock Cross-lingual transfer of named entity recognizers without parallel
  corpora.
\newblock In {\em ACL-IJCNLP}, pages 390--396, 2015.

\end{thebibliography}

\end{document}